\documentclass{article}

\usepackage[final]{styl}

\usepackage[utf8]{inputenc} 
\usepackage[T1]{fontenc}    
\usepackage{hyperref}       
\usepackage{url}            
\usepackage{booktabs}       
\usepackage{amsfonts}       
\usepackage{nicefrac}       
\usepackage{microtype}      
\usepackage{xcolor}         

\usepackage{natbib} 
\setcitestyle{authoryear} 

\usepackage{graphicx} 

\usepackage{amsmath} 
\usepackage{multirow} 

\title{Text Summarization With Graph Attention Networks}

\author{%
  Mohammadreza Ardestani\\
  Department of Computer Science\\
  University of Lethbridge\\
  Lethbridge, Alberta, Canada \\
  \texttt{ardestani@uleth.ca} \\
  \And
  Yllias Chali\\
  Department of Computer Science \\
  University of Lethbridge\\
  Lethbridge, Alberta, Canada \\
  \texttt{yllias.chali@uleth.ca} \\
}

\begin{document}

\maketitle

\begin{abstract}
This study aimed to leverage graph information, particularly Rhetorical Structure Theory (RST) and Co-reference (Coref) graphs, to enhance the performance of our baseline summarization models. Specifically, we experimented with a Graph Attention Network architecture to incorporate graph information. However, this architecture did not enhance the performance. Subsequently, we used a simple Multi-layer Perceptron architecture, which improved the results in our proposed model on our primary dataset, CNN/DM. Additionally, we annotated XSum dataset with RST graph information, establishing a benchmark for future graph-based summarization models. This secondary dataset posed multiple challenges, revealing both the merits and limitations of our models.
\end{abstract}

\section{Introduction}
\label{sec:intro}
Natural language processing (NLP) encompasses various branches, including Text-to-Text (T2T) Generation, Information Retrieval, and Machine Reasoning and Comprehension. A challenging task within T2T generation is Automatic Text Summarization, which has been extensively studied for over seven decades but remains far from achieving human-level performance \citep{Kumar2022, summarizationSurvey-waffa-el-kassas}. Early systems were predominantly extractive, but with the advent of Sequence-to-Sequence (seq2seq) models, the focus shifted towards abstractive summarization \citep{Hou2018}. 

One notable advancement in seq2seq models is the transformer \citep{vaswani2017}, originally designed for Machine Translation, which has significantly enhanced performance in various language generation tasks by generating rich intermediate representations through key, query, and value mechanisms. Despite their success in machine translation, transformers struggle with more complex tasks like Dialogue Management and Text Summarization, which require managing long dependencies and reasoning over sentence importance.

Recent findings indicate that versatile graphs, such as text graphs, syntactic graphs, semantic graphs, knowledge graphs, and hybrid graphs, can significantly enhance T2T generation tasks \citep{GNNBook-ch21-liu}. Enhancing state-of-the-art (SOTA) models with graph data structures, particularly entity-relation information graphs, has shown to improve factual consistency \cite{FAsum}. Additionally, RST \citep{rst1988} and Coref graphs can further boost summarization performance \cite{DiscoBERT}, yet there is a scarcity of graph datasets for Text Summarization. Our contribution aims to address this gap by providing a graph-annotated version of the XSum dataset \citep{xsum} and detailing the annotation process for other datasets, as well as improving our baseline models by using graph information. The dataset with complete instructions and the code of our proposed models are available on our GitHub repositories.\footnote{\href{https://github.com/Reza-Ardestani/GraphSummarizer}{https://github.com/Reza-Ardestani/GraphSummarizer}}

We have developed a stage-wise summarization model capable of processing lengthy documents and employed various optimization techniques during fine-tuning. By integrating graph structures using a Multi-layer Perceptron (MLP) architecture, we improved our baseline models' performance. We also conducted a comprehensive analysis of our results and suggested promising future directions.

\section{Related works}
\label{sec:relatedWorks}
Text summarization can be approached in three main ways: Extractive, Abstractive, and Hybrid \citep{summarizationSurvey-waffa-el-kassas}. The recent paper by \cite{DiscoBERT} introduced the Discourse-Aware Neural Extractive Text Summarization model, named DiscoBERT. Their approach segments documents into smaller units, known as Elementary Discourse Units (EDUs). They then employ two graph structures: RST and Coref graphs. These graph structures help the model understand long-range relationships within a document. They use Graph Neural Networks (GNNs) to incorporate graph information, specifically experimenting with Graph Attention Networks (GATs) \citep{gat} and Graph Convolutional Networks (GCNs) \citep{gcn}. Appendix \ref{app:GNN-main-components} elaborates more on the main components of GNNs.

There have been many abstractive approaches proposed for text summarization, one of which is BART (Bidirectional and Auto-Regressive Transformers) \citep{bart}. BART merges the strengths of BERT \citep{bert} and GPT \citep{GPT}, enabling both text understanding and generation, thus excelling in various NLP tasks. Another notable approach by \citet{FAsum} introduced an entity-relation graph to enhance summary factuality using GAT layers for node embeddings, integrating these into a transformer decoder for fact-aware generation, and a seq2seq model for post-processing factual corrections. Additionally, \citet{see2017get} proposed the Pointer-Generator Network to improve seq2seq models by generating and copying words from the source text, reducing redundancy with a coverage mechanism. Recently, \citet{BRIO} addressed the discrepancy between training and inference in abstractive summarization, suggesting a multi-task loss function to better align training with beam search-based inference, improving both training and test performance.

In recent advancements in hybrid summarization models, \cite{BertSum} proposed the BERTSumExtAbs model, combining extractive and abstractive frameworks through independent stages, extraction with BERT and abstraction with Transformers. \citet{graphRAG} enhanced retrieval-augmented generation (RAG) systems by using graph-based elements (entity nodes, relationship edges, claim covariates) and applying community detection methods to generate topic-specific summaries, which are then integrated into a global summary using Large Language Models (LLMs). \citet{pageRankForExtractiveSummary} approached long document summarization with a hybrid method, employing a graph-based extraction module that utilizes PageRank \citep{page1999pagerank} for ranking sentences based on semantic and literal similarities, combined with an encoder-decoder architecture for abstraction.

\section{Proposed model}
\label{sec:proposed model}
\begin{figure}[t]
  \includegraphics[width=\columnwidth]{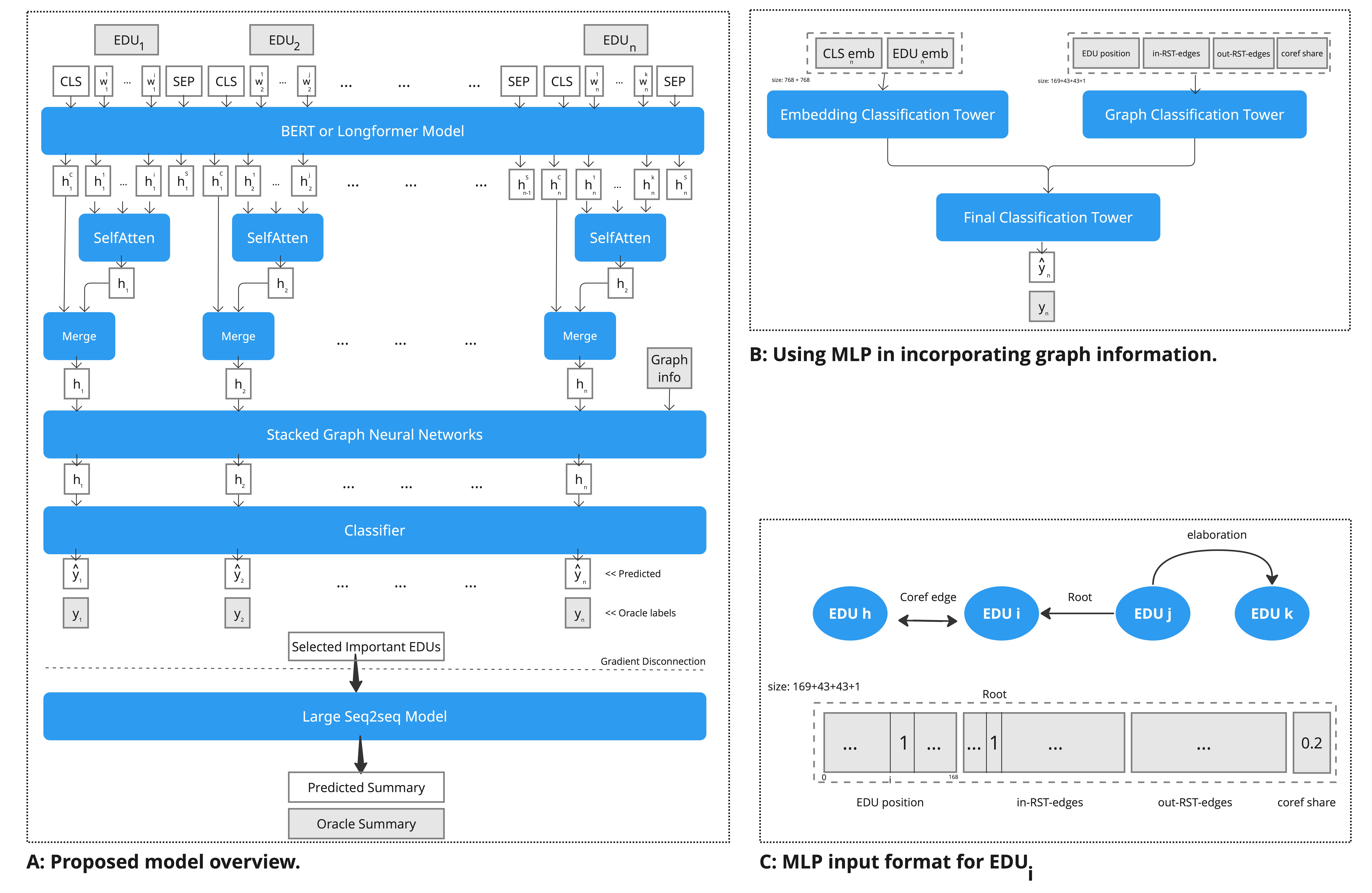}
  \caption{Proposed model overview and its MLP-based variation.}
  \label{fig:myModelOverview}
\end{figure}

This section outlines our proposed hybrid model for text summarization and its MLP variation, illustrated in Fig \ref{fig:myModelOverview}, focusing on its two stages: encoding for extraction and decoding for coherent narration. We discuss training, fine-tuning, and inference processes, and present the experimental setup for replication.

\subsection{Encoder}
The encoder classifies EDUs as important or trivial, fine-tuned using oracle labels to optimize the selection F1 score. The encoder consists of six essential steps; token addition, hidden representations, span representation, combination, GNN, and classification. First, CLS and SEP tokens are added to each EDU for classification and separation. Then, Longformer \citep{longformer} or BERT \citep{bert} are used to generate token representations. Next, the SelfAttentiveSpanExtractor aggregates token embeddings into a single embedding vector representing an EDU. These representations are then combined with CLS representations through a linear layer. Following this, GATs or MLPs are used to incorporate graph information. Finally, the combined representations are passed to a classifier to predict the importance of each EDU.

Following \citet{DiscoBERT}, we use SelfAttentiveSpanExtractor module, based on attention mechanism \citep{vaswani2017}, to aggregate contextualized token embeddings of an EDU into a single vector. This module transforms the embedding of tokens within each EDU, matrix $X$, through a linear transformation and $\text{ReLU}$ activation \citep{relu}, resulting in $H$. It then computes an attention scores, matrix $A$, for tokens within the EDU, normalizes these scores into attention weights $\alpha$ using softmax, and finally computes the span representation $S$ as the weighted sum of the token embeddings. This process, formally represented through the following equations, allows the model to emphasize relevant tokens and produce a contextually rich span representation.

\begin{equation*}
\label{eq1}
H = \text{ReLU}(XW_1 + b_1)
\end{equation*}
\begin{equation*}
\label{eq2}
A = HW_2 + b_2
\end{equation*}
\begin{equation*}
\label{eq3}
\alpha_i = \text{softmax}(A) = \frac{e^{A_i}}{\sum_{j=1}^{n}e^{A_j}}
\end{equation*}
\begin{equation*}
\label{eq4}
S = \sum_{i=1}^{n}\alpha_i X_i
\end{equation*}


Other than experiments using graphs with the GAT architecture, we have two other types of experiments. In the experiments named "without graph", we deactivated the GAT layers and did not use graph information while keeping the classifier layer in place. Secondly, we experimented with different architecture for encoding graph information than GAT. We used a MLP that takes one-hot vector representation of graph information, illustrated in the part B of Fig \ref{fig:myModelOverview}. As we are using this one-hot representations at the final layers of our model, the computational cost of it is negligible. In this figure, CLS and EDU embeddings are generated the same way as part A but we do not repeat the same layers (BERT or Longformer model and SelfAtten layers) due to space constrains. To find optimal hyperparameters of the Encoder, we ran extensive experiments on different aspects, including architecture. Later, we discovered passing EDU span embedding, CLS embedding, and graph information into one classification tower yields in better results. Coref share is the percentage of input edges for each EDU, which is only used for CNN/DM as we don't have Coref graph information for XSum dataset.

To elaborate more, we demonstrate how the graph information input vector for an EDU is generated, which is then passed to the MLP layer for predicting its importance. Figure \ref{fig:myModelOverview}, part C, illustrates the RST and Coref relationships between EDU$i$ and its neighboring nodes. There are 43 different RST relation types, including root, elaboration, comparison, and result, among others. Each EDU may have multiple incoming and outgoing RST edges. We represent this information using one-hot vector encoding, where a '1' in the vector indicates the presence of an incoming or outgoing RST edge with that specific edge type. The Coref share value represents the percentage of incoming Coref edges for each EDU. For instance, if there are a total of 10 Coref edges in the document and EDU$i$ receives two of them, its Coref share value would be 0.2. As the majority of documents have fewer than 169 EDUs, we chose 169 as the size of the EDU position vector, which makes the final dimension of the vector $2^8$.

\subsection{Decoder}
We had multiple options, such as BART or Pegasus \citep{pegasus}, for our Seq2Seq model as the decoder. Pegasus performs slightly better in R-2 and R-L metrics than BART. However, we chose BART to fine-tune, due to its fewer number of parameters. We used BART-base for CNN/DM dataset experiments and BART-large for XSum dataset. Then, we fine-tuned them on our extracted EDUs from the previous stage. As XSum dataset has fewer number of articles and significantly shorter summaries, we were able to use the large version of BART in its experiments.

\subsection{Fine-tuning and experimental setup}
During the model fine-tuning process, several key considerations were addressed. Notably, over 90\% of the EDUs were unimportant and labeled as zero, leading the model to frequently predict everything as zero. To counteract this imbalance, we increased the weight assigned to positive labels. Initially following the approach outlined in \cite{DiscoBERT}, we utilized Binary Cross Entropy (BCE) loss for our binary classification task. However, we later redefined our task within the framework of logistic regression and opted for MSE loss, which significantly improved our selection F1 score from 35\% to 39\%, approximately. This improvement was also aided by the incorporation of dropout \citep{Dropout} in the classification layer, the final segment of our encoder. This regularization technique effectively mitigated overfitting on the training set, enhanced model generalization, and thereby boosted validation dataset performance.

Another pivotal factor in our model's performance was the implementation of a learning rate scheduler. This tool was instrumental in gradually reducing the step size as the model approached the later stages of fine-tuning, which are presumptively closer to the optimal solution. Without this adjustment, the constant large steps from the initial fine-tuning phase could potentially overshoot the optimum, preventing convergence.

Our encoder model, as illustrated in the part A of Fig \ref{fig:myModelOverview}, comprises multiple components, including pre-trained language models. However, other components are not pre-trained. Therefore, we first freeze the weights of our pre-trained model and train the encoder network. Then, we unfreeze and fine-tune the entire encoder network for four epochs. This approach gives the layers after the language model, which are initially randomly initialized, more time to adjust themselves.

Our primary computational resources were Cedar and Narval servers from Compute Canada\footnote{https://docs.alliancecan.ca/.}, which are equipped with V100, and A100 GPUs. To accelerate the fine-tuning process, we employed the Distributed Data Parallel (DDP) feature from PyTorch\footnote{https://pytorch.org/docs/.} following the configuration guidelines provided on the Compute Canada data parallelization webpage\footnote{https://docs.alliancecan.ca/wiki/PyTorch.}. This setup allowed us to utilize 4 GPUs simultaneously, significantly enhancing our fine-tuning efficiency. Full details, scripts, code, and utilities of our research are available on our GitHub repository.

\section{Evaluation}
\label{sec:eval}
In this section, we will present the results of our experiments, and the effectiveness of our proposed approach.

\subsection{Results}

Table \ref{tab:encoder-result-CNNDM} and Table \ref{tab:encoder-result-XSum} demonstrate the performance of our different encoder models on CNNDM and XSum datasets. The first group of experiments uses Longformer, while the second uses BERT. For BERT, input documents were truncated to a maximum of 512 tokens. Our primary criterion for evaluating the optimal encoder module was based on the F1 selection score. Although Longformer-based models generally achieve lower F1 score than their BERT-based counterparts, they have a window size that is almost three times longer.

\begin{table}[ht]
\footnotesize
\caption{Test results of the encoder models on CNNDM dataset.}
\label{tab:encoder-result-CNNDM}
\centering
\begin{tabular}{llccc}
\hline
& \textbf{Model} & \textbf{Precision} & \textbf{Recall} & \textbf{F1} \\ \hline
\multirow{5}{*}{\rotatebox{90}{Longformer}} & without graph & 28.98 & 56.94 & \underline{38.41} \\
& w. GAT for RST  & 28.75 & 54.58 & 37.66 \\
& w. GAT for Coref & 26.57 & 57.26 & 36.30 \\
& w. GAT for RST \& Coref & 27.62 & 57.94 & 37.40 \\
& w. MLP for RST \& Coref & 29.81 & 54.94 & \textbf{38.65} \\ \hline
\multirow{5}{*}{\rotatebox{90}{BERT}} & without graph & 28.61 & 65.00 & 39.74 \\
& w. GAT for RST & 29.63 & 61.64 & 40.02 \\
& w. GAT for Coref & 29.63 & 61.84 & 40.06 \\
& w. GAT for RST \& Coref & 29.74 & 61.89 & \textbf{40.18} \\
& w. MLP for RST \& Coref & 29.64 & 62.17 & \underline{40.15} \\ \hline
\end{tabular}
\end{table}

\begin{table}[ht]
\caption{Test results of the encoder models on XSum dataset.}
\label{tab:encoder-result-XSum}
\centering
\footnotesize
\begin{tabular}{llccc}
\hline
& \textbf{Model} & \textbf{Precision} & \textbf{Recall} & \textbf{F1} \\ \hline
\multirow{5}{*}{\rotatebox{90}{Longformer}}\\ & without graph & 30.71 & 50.04 & \underline{38.06} \\
& w. GAT for RST  & 30.89 & 48.06 & 37.61 \\
& w. MLP for RST & 30.86 & 49.74 & \textbf{38.09} \\\\ \hline
\multirow{3}{*}{\rotatebox{90}{BERT}} & without graph & 28.98 & 56.94 & \underline{38.41} \\
& w. GAT for RST  & 28.75 & 54.58 & 37.66 \\
& w. MLP for RST & 29.81 & 54.94 & \textbf{38.65} \\ \hline
\end{tabular}
\end{table}

We use our fine-tuned encoder modules to predict important EDUs which will be fed into the final stage, which uses BART model. Upon applying the decoder module over the concatenated EDUs, we obtain the final summary generated by our hybrid approach. Table \ref{tab:EncoderDecoder-result-CNNDM} and Table \ref{tab:EncoderDecoder-result-XSum} depict the evaluation result of our generated summaries which can be compared with previous models as our baseline. Average ROUGE \citep{rouge} scores are obtained by 1000 replicates and ± is followed by an estimated margin of error under 95\% confidence \citep{koehn-2004-statistical} and the margin of error is calculated as the average difference between the lower and upper
bounds of the scores. The minimal impact of using graphs in BERT experiments suggests that graphs are more effective in longer document summarizations.

\begin{table}[ht]
\caption{Test results of the encoder-decoder models on CNNDM dataset.}
\label{tab:EncoderDecoder-result-CNNDM}
\centering
\footnotesize
\begin{tabular}{llccc}
\hline
& \textbf{Model} & \textbf{R-1} & \textbf{R-2} & \textbf{R-L} \\ \hline
\multirow{5}{*}{\rotatebox{90}{Longformer}} &without graph & \underline{43.60} ± 0.23 & \underline{20.77} ± 0.25 & \underline{40.60} ± 0.22 \\
& w. GAT for RST  & 43.01 ± 0.23 & 20.35 ± 0.25 & 40.00 ± 0.21 \\
& w. GAT for Coref & 43.04 ± 0.22 & 20.32 ± 0.25 & 39.98 ± 0.24 \\
& w. GAT for RST \& Coref & 43.30 ± 0.23 & 20.55 ± 0.23 & 40.30 ± 0.23\\
& w. MLP for RST \& Coref & \textbf{43.73} ± 0.22 & \textbf{20.91} ± 0.26 & \textbf{40.72} ± 0.24 \\ \hline
\multirow{5}{*}{\rotatebox{90}{BERT}} & without graph  & \textbf{43.43} ± 0.24 & 20.62 ± 0.24 & \textbf{40.40} ± 0.23  \\
& w. GAT for RST  & 43.37 ± 0.22 & \underline{20.65} ± 0.25 & 40.34 ± 0.23 \\
& w. GAT for Coref  & 43.40 ± 0.23 & 20.64 ± 0.24 & 40.39 ± 0.21 \\
& w. GAT for RST \& Coref & 43.34 ± 0.23 & 20.64 ± 0.25 & 40.34 ± 0.24  \\
& w. MLP for RST \& Coref & \underline{43.42} ± 0.23 & \textbf{20.70} ± 0.25 & \underline{40.34} ± 0.23  \\ \hline
\end{tabular}
\end{table}

\begin{table}[ht]
\caption{Test results of encoder-decoder models on XSum dataset.}
\label{tab:EncoderDecoder-result-XSum}
\centering
\footnotesize
\begin{tabular}{llccc}
\hline
& \textbf{Model} & \textbf{R-1} & \textbf{R-2} & \textbf{R-L} \\ \hline
\multirow{5}{*}{\rotatebox{90}{Longformer}}\\ & without graph & \underline{36.56} ± 0.25 & \underline{14.28} ± 0.23 & \underline{28.29} ± 0.23  \\
& w. GAT for RST  & 36.24 ± 0.26 & 14.14 ± 0.23 & 28.14 ± 0.24\\
& w. MLP for RST  & \textbf{36.82} ± 0.27 & \textbf{14.66} ± 0.24 & \textbf{28.67} ± 0.26  \\\\ \hline
\multirow{3}{*}{\rotatebox{90}{BERT}} & without graph & \textbf{36.41} ± 0.28 & \textbf{14.23} ± 0.25 & \textbf{28.37} ± 0.27  \\
& w. GAT for RST & 35.33 ± 0.25 & 13.30 ± 0.23 & 27.45 ± 0.25\\
& {w. MLP for RST}  & \underline{35.97} ± 0.25 & \underline{13.95} ± 0.24 & \underline{27.90} ± 0.26 \\ \hline
\end{tabular}

\end{table}

Our best model on CNNDM dataset is compared with previous models as our baseline in Table \ref{tab:CNNDM_res}. The term "Lead3" refers to the first three sentences of the input document. Remarkably, by simply selecting Lead3, the results approach the performance of neural network based models. This observation underscores the extractive nature of the CNNDM dataset. In this table, we present three baseline models that closely resemble our proposed model, spanning three different categories of summarization systems.

\begin{table}[ht]
\caption{Comparing our best model with previous models as our baseline on CNNDM.}
\label{tab:CNNDM_res}
\centering
\footnotesize
\begin{tabular}{lccc}
\hline
\textbf{Model} & \textbf{R-1} & \textbf{R-2} & \textbf{R-L} \\ 
\hline
\multicolumn{4}{c}{Oracle} \\
\hline
Lead3 & 40.42 & 17.62 & 36.67 \\
Oracle (EDUs)  & 61.61 & 37.82 & 59.27 \\
\hline
\multicolumn{4}{c}{Extractive} \\
\hline
DiscoBERT \citep{DiscoBERT} & 43.38 & 20.44 &  40.21 \\
DiscoBERT w Coref  & 43.58 & 20.64 &  40.42 \\
DiscoBERT w RST   & 43.68 & 20.71 &  40.54 \\
DiscoBERT w RST \& Coref & \textbf{43.77} & \underline{20.85} &  \underline{40.67} \\
BERTSUMEXT \citep{BertSum} & 43.25 & 20.24 & 39.63 \\
\hline
\multicolumn{4}{c}{Abstractive} \\
\hline
BERTSUMABS \citep{BertSum} & 41.72 & 19.39 & 38.76 \\
BART-base\(^{\dagger}\) \citep{bart} & 41.44 & 18.49 & 38.32 \\
\hline
\multicolumn{4}{c}{Hybrid} \\
\hline
BERTSUMEXTABS \citep{BertSum} & 42.13 & 19.60 & 39.18 \\
\textbf{Longformer with MLP for RST \& Coref (ours)}  & \underline{43.73} & \textbf{20.91} & \textbf{40.72} \\
\hline
\end{tabular}
\begin{minipage}{0.72\textwidth}
\(^{\dagger}\) We generated the result based on their provided model and hyperparameters on our CNNDM testset. 
\end{minipage}
\end{table}

\subsection{Discussions}

In this section, we delve deeper into our findings, examining them across multiple key aspects. Our main goal for this study was to improve the baseline models' results while thoroughly testing our models. This led us to select a secondary challenging dataset with multiple contrasting attributes compared to CNNDM. The XSum dataset, being highly abstractive and concise, allowed us to better assess our approach compared to other studies that might choose similar secondary datasets to reconfirm their initial results.


Regarding the use of graphs, we found that leveraging RST and Coref graphs improved the "without graph" version and, in the case of CNNDM, our approach outperformed our baseline models. The method of incorporating graph information is critical, however. We operated our graph-based encoders under nearly identical settings. Despite various attempts, meticulous hyperparameter tuning, and significant resource allocation to GAT-based experiments, none achieved higher ROUGE \citep{rouge} scores in their final summaries. Testing GNNs with edge labels also did not improve the performance. This suggests that the prevailing approach of using GNNs for incorporating graph information might not always be suitable, and a simple MLP could be more effective.   


Encoder models reported in Table \ref{tab:encoder-result-CNNDM} and Table \ref{tab:encoder-result-XSum} are not the only ones we experimented with. For each model, we tested multiple combinations of hyperparameters and design choices. For example, the initial predecessor of the 'Longformer without GAT' model achieved only a 35\% F1 score until we optimized the learning rate scheduler, dropout layers, loss function, and optimizer weight. A notable technique involved assigning greater weight to positive labels. Given that approximately 10\% of EDUs are labeled important in both the CNNDM and XSum datasets, this approach addressed issues of slow convergence and overfitting in our imbalanced datasets.


We approached the fine-tuning of encoder models differently. For models using GAT, we fine-tuned them in two phases. First, we froze the weights of the language model and trained all other layers. Then, we unfroze the language model weights and fine-tuned the entire network for a few epochs. However, this approach did not improve GAT performance. Moreover, while most studies use only ROUGE scores to evaluate performance, we employed several additional evaluation metrics (see Appendix \ref{app:Additional-Evaluation}) to facilitate comparison of our best model with future models. Regarding the language models in our encoders, we found that Longformer-based models generally performed better than their BERT-based counterparts.


Regarding our extraction-abstraction (hybrid) approach, our analysis indicates that an extraction approach based on greedy labeling might not be appropriate. Upon applying BART model in zero-shot setting on oracle EDUs of CNNDM, we reached an extremely high ROUGE score than BART itself. However, this improvement was not observed when using the fine-tuned BART-large model on oracle EDUs from the XSum dataset, as detailed in Table \ref{tab:XSum_res} and numerically analyzed in Table \ref{tab:xsum_cnndm_extractive_approaches.}. This discrepancy highlights the potential limitations of the extraction approach and greedy labeling. This discrepancy underscores the need for further investigation into the effectiveness of different strategies across various models and datasets. It also raises questions about the adaptability and generalizability of these approaches, suggesting that what works well for one dataset or model might not necessarily translate to success with another.

\begin{table}[ht]
\caption{Comparing our best model with baseline models on XSum.}
\footnotesize
\centering
\begin{tabular}{lccc}
\hline
\textbf{Model} & \textbf{R-1} & \textbf{R-2} & \textbf{R-L} \\ 
\hline
\multicolumn{4}{c}{Oracle} \\
\hline
LEAD1 \(^{\dagger}\) & 16.30 & 1.60  & 11.95 \\
Oracle Sentence \citep{BertSum} & 29.79 & 8.81 & 22.66 \\ 
Oracle EDU (ours) & 36.03 & 11.47 & 30.86 \\ 
\hline
\multicolumn{4}{c}{Abstractive} \\
\hline
BERTSUMABS \citep{BertSum} & 38.76 & 16.33 & 31.15 \\
BART-large \(^{\ddagger}\) \citep{bart} & \textbf{42.34} & \textbf{18.45} & \textbf{32.40} \\
Zeroshot BART-large on oracle EDUs  & 31.94 & 11.10 & 24.62 \\
Finetuned BART-large on oracle EDUs & 42.34 & 18.44 & 32.40 \\
\hline
\multicolumn{4}{c}{Hybrid} \\
\hline
BERTSUMEXTABS \citep{BertSum} & 38.81 & 16.50 & 31.27 \\
Longformer with MLP for RST (ours) & 36.82 & 14.66 & 28.67 \\
\hline
\end{tabular}
\begin{minipage}{0.67\textwidth}
\(^{\dagger}\) The term "Lead1" refers to the first sentences of the input documents. \(^{\ddagger}\) We generated this result based on their provided fine-tuned model.
\end{minipage}
\label{tab:XSum_res}

\end{table}

\begin{table}[ht]
\caption{Comparing the performance of extractive with abstractive approaches on CNNDM and XSum.}
\footnotesize
\centering
\begin{tabular}{lccc}
\hline
\textbf{Model} & \textbf{R-1} & \textbf{R-2} & \textbf{R-L} \\ 
\hline
\multicolumn{4}{c}{CNNDM} \\
\hline
BART-base  & 41.44 & 18.49 & 38.32 \\
Oracle (EDUs)  & \textbf{61.61} ($\uparrow$) & \textbf{37.82} ($\uparrow$) & \textbf{59.27} ($\uparrow$) \\
Zeroshot BART on oracle EDUs  & 53.43 ($\uparrow$) & 32.20 ($\uparrow$) & 51.05 ($\uparrow$) \\
Finetuned BART on oracle EDUs & 60.89 ($\uparrow$) & 37.59 ($\uparrow$) & 57.47 ($\uparrow$) \\
\hline
\multicolumn{4}{c}{XSum} \\
\hline
BART-large & \textbf{42.34} & \textbf{18.45} & \textbf{32.40} \\
Oracle (EDUs) (ours) & 36.03 ($\downarrow$) & 11.47 ($\downarrow$) & 30.86 ($\downarrow$)\\ 
Zeroshot BART on oracle EDUs  & 31.94 ($\downarrow$) & 11.10 ($\downarrow$) & 24.62 ($\downarrow$) \\
Finetuned BART on oracle EDUs & 42.34 ($\downarrow$) & 18.44 ($\downarrow$) & 32.40 ($\downarrow$) \\
\hline
\end{tabular}

\label{tab:xsum_cnndm_extractive_approaches.}
\end{table}

\section{Conclusion and future directions}
\label{sec:conclusion}

In conclusion, we have introduced a hybrid summarization model which uses BERT as well as Longformer. We, initially, intended to use GAT architecture to incorporate graph information. Despite many hyperparameter search or testing multiple architectural changes, GAT did not improved the performance in our settings. Later, we used an MLP for incorporating graph information which increased the performance. On CNNDM dataset, the MLP version of our model surpasses previous baseline approaches, including extractive, abstractive, and hybrid models, demonstrating superior performance.

In our work, we provided an annotated version of the XSum dataset, refer to Appendix \ref{app:Dataset-Generation}, enhanced with RST graphs and higher oracle ROUGE scores. Our detailed data processing instructions are designed to facilitate further research, offering a roadmap for other researchers to apply these methodologies to other datasets. 

Looking forward, we identify promising directions for future research, both in terms of data processing techniques and model architecture enhancements. In terms of architecture, a promising approach is to address the gradient disconnection issue by using an end-to-end model for summarization. Inspired by the SimpleTOD model in dialogue systems \citep{SimpleTOD}, we can cast the extraction and summarization into one module by passing the input document, RST-graph, and other graph information to a single language model to generate important EDU indices, and subsequently, the final summaries. For data processing, three key directions are identified: employing semantic-aware metrics like BERTscore \citep{bertscore} instead of syntax-sensitive ones like ROUGE, generating Coref graphs for datasets like XSum, and creating graphs for longer datasets to improve document representation. These advancements can significantly enhance the efficacy of document summarization models.

\bibliography{custom}







\appendix

\section{GNN main components}
\label{app:GNN-main-components}

\begin{figure}[t]
  \includegraphics[width=\columnwidth]{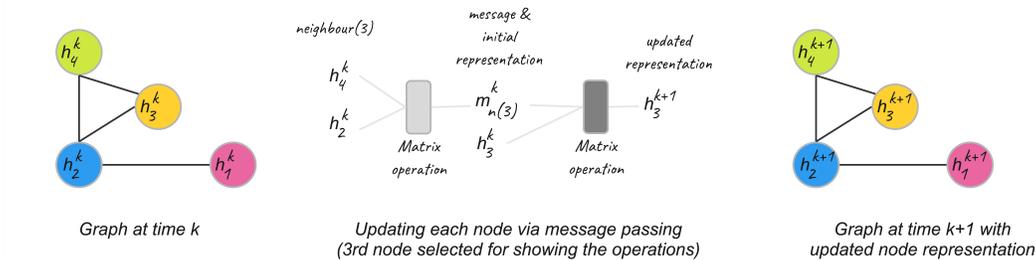}
  \caption{Message aggregation and update in GNNs for a single node from its local
neighborhood.}
  \label{fig:gnn}
\end{figure}

We will briefly elaborate on GNNs. Graphs consist of nodes and edges, each with features, and the main task in graph learning is to generate rich node embeddings for predictions. GNNs achieve this through two operations: message aggregation and update \citep{Hamilton-William-graphs} and these operations are illustrated in Fig \ref{fig:gnn}. During each iteration of the message-passing phase in a GNN, the hidden embedding \(h_u^{(k)}\) for each node \(u \in V\) is updated based on information aggregated from \(u\)'s neighborhood, \(N(u)\). This process is expressed as follows:

\begin{align*}
    m^{(k)}_{N(u)} &= \text{AGGREGATE}^{(k)} \left(\{h_v^{(k)}, \forall v \in N(u)\}\right)
\end{align*}
\begin{align*}
    h_u^{(k+1)} &= \text{UPDATE}^{(k)}\left(h_u^{(k)}, m^{(k)}_{N(u)}\right)
\end{align*}

\texttt{UPDATE} and \texttt{AGGREGATE} are differentiable functions, such as neural networks, and \(m_{N(u)}\) denotes the aggregated message from \(N(u)\). Superscripts differentiate embeddings and functions across message-passing iterations, often referred to as different layers of the GNN, which may not share weights \citep{Hamilton-William-graphs}.

\section{Additional evaluation}
\label{app:Additional-Evaluation}
We evaluate the quality of generated summaries using additional metrics. Table \ref{tab:Novel-N-grams-Proportions} compares the proportions of novel N-grams across different text types. Human-generated (Reference) summaries demonstrate the highest novelty, followed by predicted summaries using our decoder generator, while Oracle EDUs and Selected EDUs show no novelty in 1-grams but some in bigrams and trigrams due to novel segment concatenations. Further, we report BERT-score \citep{bertscore} and BART-score \citep{bartscore} for our best model in Table \ref{sample-metrics-table2}, providing benchmarks for future comparisons, as BLUE score was deemed irrelevant for our task.

\begin{table}[ht]
\centering
\caption{Novel n-grams proportions of our best model on CNN/DM}
\label{tab:Novel-N-grams-Proportions}
\begin{tabular}{lccc}
\hline
Text & n=1 & n=2 & n=3 \\ \hline
Oracle EDUs           & 0.000 & 0.015 & 0.071 \\
Selected EDUs           & 0.000 & 0.012 & 0.053 \\
Predicted Summaries           & 0.005 & 0.060 & 0.171 \\
Reference Summaries              & 0.008 & 0.173 & 0.491 \\ \hline
\end{tabular}
\end{table}

\begin{table}[ht]
\centering
\caption{BERT and BART scores of our best model on CNN/DM.}
\label{sample-metrics-table2}
\begin{tabular}{lcc} 
\hline
Model & BERT Score & BART Score \\ \hline
Our Best Model & 0.87 & -3.73 \\ \hline
\end{tabular}
\end{table}

\section{Dataset generation}
\label{app:Dataset-Generation}

\begin{figure}[ht]
    \centering
    \includegraphics[width=0.8\linewidth]{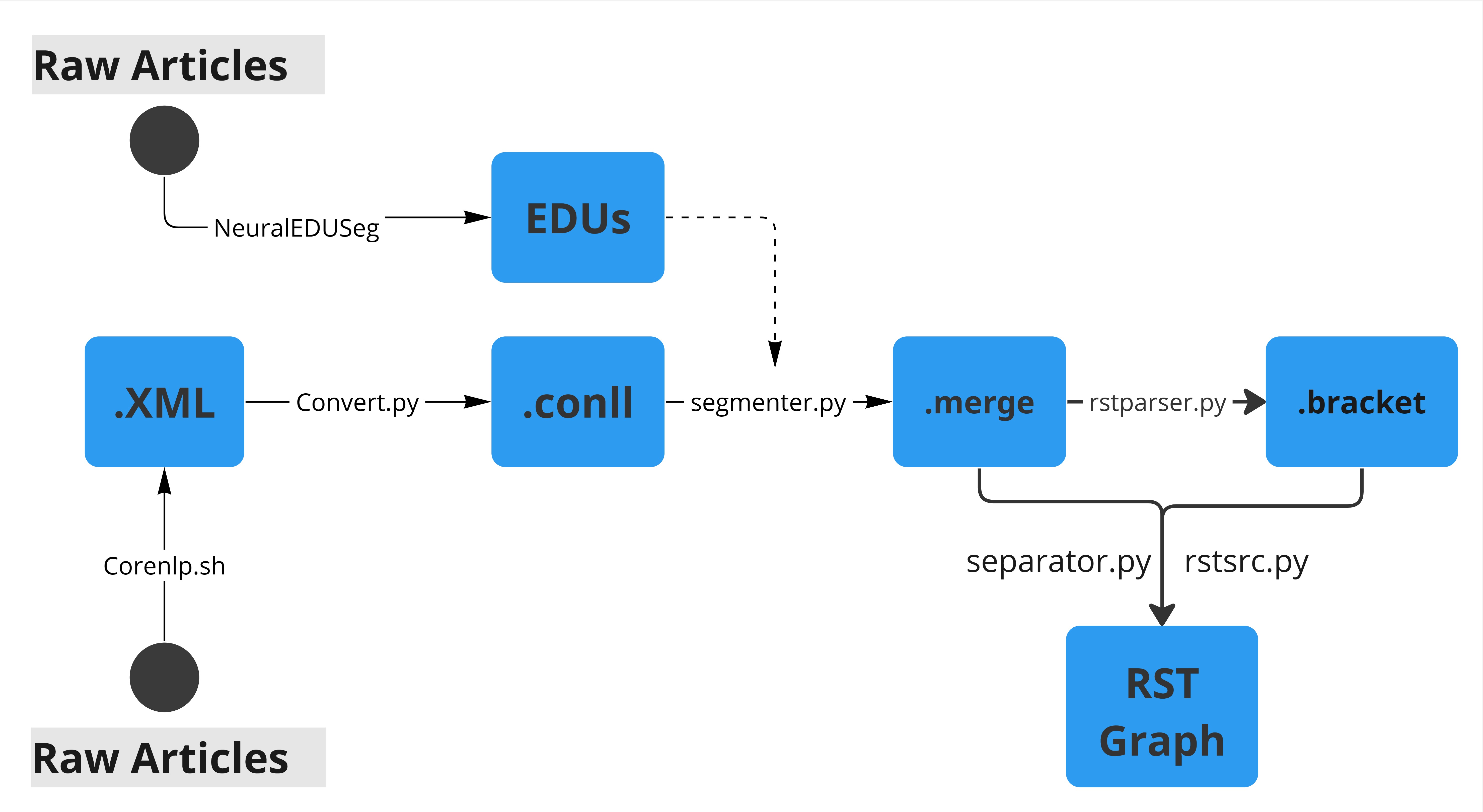}
    \caption{XSum data processing pipeline.}
    \label{img:pipeline}
\end{figure}

Using the pipeline depicted in Fig \ref{img:pipeline}, our objective is to generate EDUs, assign labels to these EDUs, and construct the RST graph for each raw XSum document. There are some byproducts in this process, namely RST tree or .XML files, that we only discuss briefly.

To begin with, we start from the lower left of Fig \ref{img:pipeline} which uses Stanford CoreNLP\footnote{\href{https://github.com/stanfordnlp/CoreNLP}{https://github.com/stanfordnlp/CoreNLP.}} \citep{StanfordCoreNLP} to transform raw documents into .XML files which contains nine attributes for each token in the raw document. Attributes are: sentence index, token index (within a sentence), token, lemma, POS
(Part Of Speech tags), dependency label, dependency head, NER (Named Entity Recognition), and partial constituent parse. 

Using the code of a recent study, DPLP \citep{DPLP}, we generate .conll files that are a simplified format of .XML files. They contain the same information as .XML files, however. Then, from .conll files we generate .merge files that has the same information as .conll file with another attribute, that is the token's EDU index. In fact, EDU segmentation is done at this stage. Moving forward we generate .bracket files that contain RST tree of documents. Lastly, from both .merge and .bracket files we generate RST graphs of the documents. Now, we have both EDUs and RST graphs. To generate the label of EDUs, we use the previously discussed algorithm.

There is also one other arrow joining the .conll to .merge phase. This is because the aforementioned generated EDUs, by DPLP, are not highly accurate. So, we use another recent study, NeuralEDUseg \citep{NeuralEDUseg} to generate the EDU index of tokens and use this result to update the original .merge files.

For fine-tuning the extraction part of our proposed model, we need oracle labels for EDUs. We, inspired by \citep{DiscoBERT}, implemented a greedy selection function takes the list of EDUs of a document, the list of EDUs from the oracle summary, and the desired maximum size of the summary. We iteratively select EDUs that, when added to the current selection, maximally increase the overall ROUGE score of the selection with respect to the oracle summary. Now, the processed documents are ready to be used in our graph-based approach. Table \ref{table:processed_xsum_stat} present the statistics of this annotated dataset. Our described approach occasionally generates an empty RST graph and an empty list of oracle labels due to the shortness and extreme abstractness of XSum articles, respectively.
\begin{table}[ht]
\caption{Statistics of the Processed XSum.}
\centering
\begin{tabular}{lrrrrrr}
\toprule
       Dataset & \multicolumn{2}{c}{Average} & \multicolumn{3}{c}{Empty} &  Total Records \\
       & EDUs & Summary & RST Graph & Oracle Label & Both & \\
\midrule
Train &                   52.35 &                       21.10 &          121 &         1944 &         102 &         204017 \\
Test &                   41.85 &                       21.10 &            7 &           85 &           5 &          11333 \\
Valid &                   41.39 &                       21.13 &           10 &           96 &           8 &          11327 \\
\bottomrule
\end{tabular}
\begin{minipage}{0.88\textwidth}
\textit{Note: Average column shows average number of EDUs in articles and average number of words in summaries of articles. Empty indicates how many documents have empty RST graph, oracle labels, or both.}
\end{minipage}
\label{table:processed_xsum_stat}
\end{table}

\section{Architectural details for reproducibility}
\label{app:reproducibility} 

We have provided an overview of the model and elaborated on key components, including specifications of the pre-trained language models used in both the encoder and decoder blocks. In this section, we delve into the specifics of the GNN module and the classification module, as illustrated in Fig \ref{fig:myModelOverview}.

\begin{figure}[ht]
    \centering
    \includegraphics[width=0.4\linewidth]{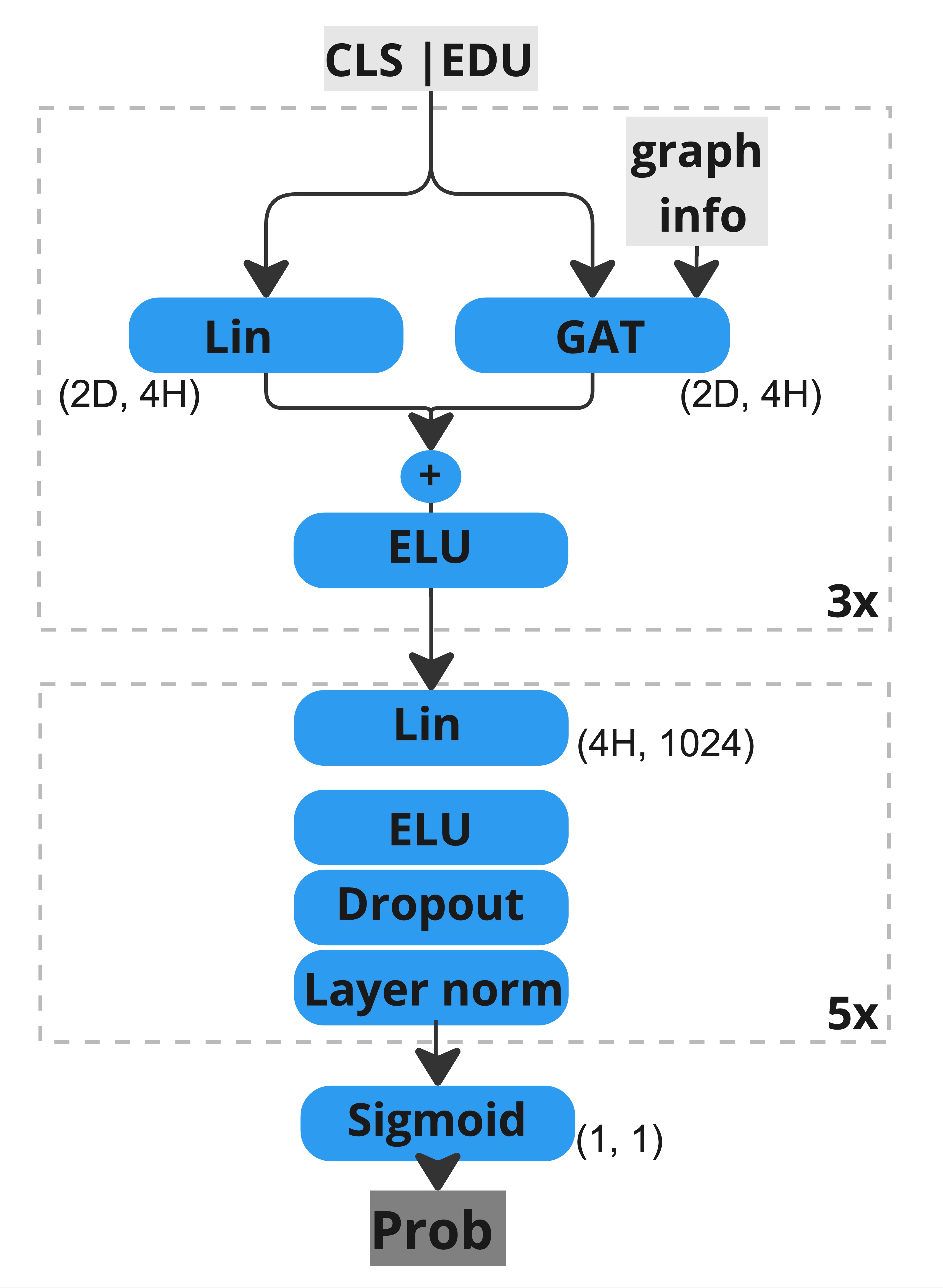}
    \caption{Details of GNN and classification modules.}
    \label{img:gat_classification}
\end{figure}

For the stacked GNN module, we leveraged the torch-geometric library\footnote{Website: https://pytorch-geometric.readthedocs.io/en/latest/}, a specialized framework for working with graph-structured data. The GNN module receives merged CLS and EDU embeddings, each with a dimension of 768 (denoted as D in Fig \ref{img:gat_classification}), along with graph-based information, such as RST structure. This module consists of a linear layer and a GAT layer with four attention heads and a hidden dimension of 256 (denoted as H). After applying summation and an Exponential Linear Unit (ELU) \citep{elu} activation function, the output is passed through two additional GAT and linear layers with the same architecture. 
For experiments utilizing both RST and Coref information, we concatenate the outputs of their respective GNN blocks, where each block has its own set of weights and does not share parameters. The concatenated representation is then passed into the classification module for further processing.

The classification module begins with a linear layer whose input and output dimensions are detailed in Fig \ref{img:gat_classification}. This layer is followed by an activation function to introduce non-linearity, a dropout layer \citep{Dropout} to mitigate overfitting, and a normalization layer \citep{layernorm} for training stability. The same sequence of layers is repeated four more times with output dimensions of 1024, 1024, 64, and 1, respectively, gradually reducing the feature representation before the final classification.

\section{Assessment of MLP-GAT performance gap}
\label{app:performance-gap}

While theoretical proofs and explanations of large neural model behavior remain challenging, establishing empirical analysis and systematic validation is paramount. In this section, we firstly analyze the information encoded within graph structures and secondly investigate the key factors driving performance variations across our different graph-incorporating architectures.

\begin{figure}[ht]
    \centering
    \includegraphics[width=0.98\linewidth]{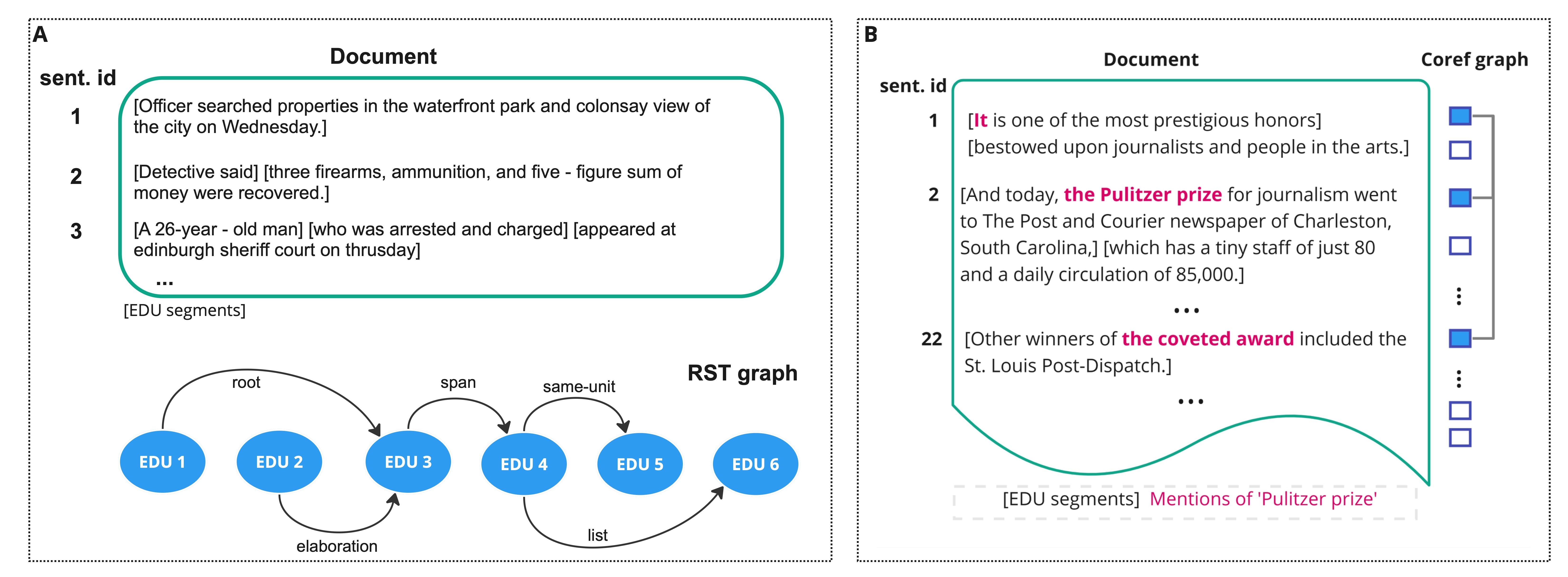}
    \caption{RST and Coref graph examples. Document A is selected from the test set of our processed XSum dataset. Part B, adapted from \citep{DiscoBERT}, shows Coref edges for a particular named entity.}
    \label{img:rst-coref-example}
\end{figure}

To better demonstrate the salient information captured by different graph structures, we present two examples in Fig \ref{img:rst-coref-example}. RST graphs capture the rhetorical dependencies between different EDUs within a document, as well as the roles these units play in the overall discourse. Similarly, Coref graphs trace reference dependencies between EDUs, storing these relations in bidirectional graphs which does not have any edge type. This graph-based information can significantly aid manual summarization efforts. Moreover, in cases where language models have not been specifically trained to extract or leverage such information, incorporating these graphs into the summarization process is likely to enhance the performance of automatic summarization models.

\cite{DiscoBERT} argue that the BERT model struggles to capture long-range dependencies. To address this, they incorporate RST and Coref graph structures to update embeddings using the GAT architecture. However, since the GAT architecture cannot handle edge types, they omit the use of RST-edge types. In contrast, we found that more recent models, such as Longformer, do not face the same limitation as BERT, thanks to their sliding-window attention mechanism and diverse pre-training tasks. However, we observed that RST edge types for EDUs and the ratio of Coref input edges to EDUs are informative. We incorporated this information using a simple MLP, which we found to be sufficient. Additionally, we experimented with Relational graph convolution networks \citep{relationalgraph} using the DGL library\footnote{https://docs.dgl.ai/en/2.0.x/generated/dgl.nn.pytorch.conv.RelGraphConv.html.} to integrate both RST and Coref graph information, including RST edge types. However, this approach did not yield better results to report.


\end{document}